%% file: main.tex
\documentclass[twocolumn]{svjour3}          

\smartqed  

\usepackage{natbib}
\usepackage{lastpage}
\PassOptionsToPackage{hyphens}{url}\usepackage{hyperref}
\hypersetup{
  colorlinks,%
  citecolor=blue,%
  filecolor=blue,%
  linkcolor=blue,%
  urlcolor=blue
}
\usepackage{cite}
\usepackage{adjustbox}
\usepackage{multirow}
\usepackage{enumitem}

\renewcommand{\cite}[1]{\citep{#1}}
\usepackage{mathrsfs}
\usepackage{graphicx}
\usepackage{amsmath}
\usepackage{amsfonts}

\usepackage{pifont}
\newcommand{\cmark}{\ding{51}}%
\newcommand{\xmark}{\ding{55}}%

\newcommand{\na}{n/a}

\journalname{International Journal of Computer Vision}
\begin{document}

\title{Neural Architecture Search for Dense Prediction Tasks in Computer Vision}

\titlerunning{NAS for Dense Prediction Tasks}        


\author{Thomas Elsken$^{1}$ \and Arber Zela$^{2}$ \and Jan Hendrik Metzen$^{1}$ \and Benedikt Staffler$^{1}$ \and Thomas Brox$^{2}$ \and Abhinav Valada$^{2}$ \and Frank Hutter$^{2,1}$} 

\authorrunning{Elsken et al.} 

\institute{Thomas Elsken \at
          {thomas.elsken@de.bosch.com}
          \and
          Arber Zela \at
          {zelaa@cs.uni-freiburg.de}
          \and 
          Jan Hendrik Metzen \at
          {janhendrik.metzen@de.bosch.com}
          \and 
          Benedikt Staffler \at
          {benediktsebastian.staffler@de.bosch.com}
          \and
          Thomas Brox \at
          {brox@cs.uni-freiburg.de}
          \and
          Abhinav Valada \at
          {valada@cs.uni-freiburg.de}
          \and
          Frank Hutter \at
          {fh@cs.uni-freiburg.de}
          \and
          {
          $^{1}$ Bosch Center for Artificial Intelligence, Robert Bosch GmbH, Germany.\\
          $^{2}$ University of Freiburg, Germany.}
}

\date{ }

\maketitle
\begin{abstract}
The success of deep learning in recent years has lead to a rising demand for neural network architecture engineering. As a consequence, neural architecture search (NAS), which aims at automatically designing neural network architectures in a data-driven manner rather than manually, has evolved as a popular field of research. With the advent of weight sharing strategies across architectures, NAS has become applicable to a much wider range of problems. In particular, there are now many publications for dense prediction tasks in computer vision that require pixel-level predictions, such as semantic segmentation or object detection. 
These tasks come with novel challenges, such as higher memory footprints due to high-resolution data, learning multi-scale representations, longer training times, and more complex and larger neural architectures.
In this manuscript, we provide an overview of NAS for dense prediction tasks by elaborating on these novel challenges and surveying ways to address them to ease future research and application of existing methods to novel problems.
    
\end{abstract}

\input{chapters/introduction}

\input{chapters/recap_and_nas_for_dpt}

\input{chapters/segmentation}

\input{chapters/object_det}

\input{chapters/ongoing_future_work}

\acknowledgement{Robert Bosch GmbH is acknowledged for financial support.}

\bibliographystyle{spbasic}      

\begin{small}
\bibliography{nas_cv}
\end{small}

\end{document}

%% file: chapters/introduction.tex
\section{Introduction}

With the advent of deep learning, features are no longer manually designed but rather learned in an end-to-end fashion from data, resulting in impressive results for various problems, such as image recognition~\citep{NIPS2012_4824}, speech recognition~\citep{38131}, machine translation~\citep{DBLP:journals/corr/BahdanauCB14}, or reasoning in games~\citep{alpha_go}. This, however, lead to a new design problem: the feature engineering process is replaced by engineering neural network architectures~\citep{DBLP:journals/corr/SimonyanZ14a,He16,Szeg16,Szegedy:2017:III:3298023.3298188,howard_mobilenets:_2017,NIPS2014_5423,Zhang_2018_CVPR,long_fcnet,6909475,7410526,NIPS2015_5638,7780460,10.1007/978-3-319-46448-0_2,unet,pmlr-v97-tan19a,mohan2020efficientps,Cheng_2020_CVPR,Zhong_2020_CVPR}.
This architectural engineering is especially prevalent for dense prediction tasks in computer vision, such as semantic segmentation, object detection, optical flow estimation, or disparity estimation. These tasks typically require complex neural architectures, often composed of various components, each having a different purpose, e.g., extracting features at different scales, feature fusion across levels, or dedicated architectural heads for, e.g., generating bounding boxes or making class predictions.

\begin{figure*}
		\centering
\includegraphics[width=\linewidth]{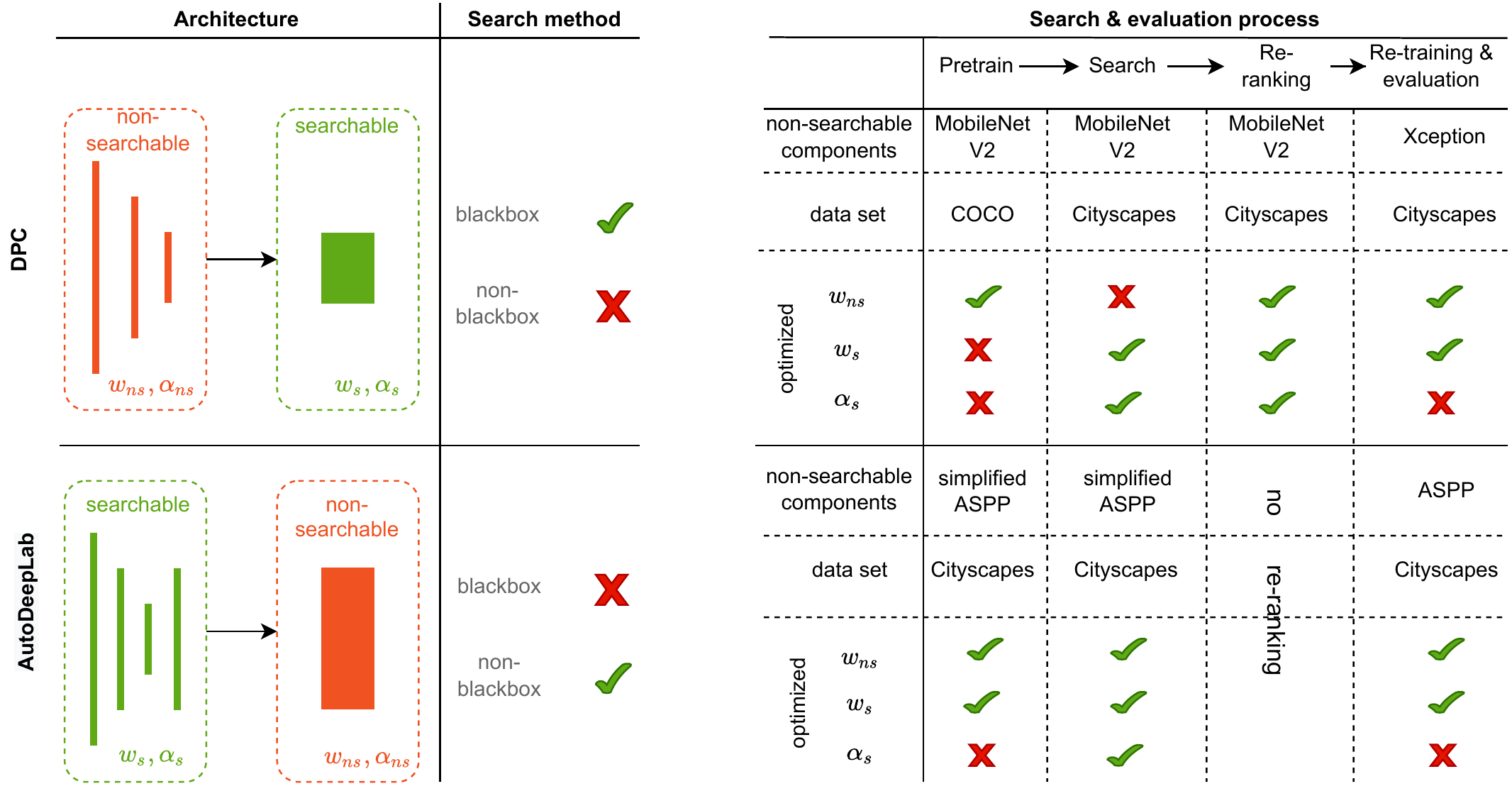}	
		\caption{
		Visualization of the widely differing architecture search process for Auto-DeepLab~\citep{Liu_2019_CVPR} and dense prediction cell (DPC)\citep{NIPS2018_8087}. 
		Left: illustration of the overall architecture and which components of the architecture are searchable. \citet{NIPS2018_8087} fix the encoder and search for a dense prediction cell to encode multi-scale information, while \citet{Liu_2019_CVPR} search for the encoder and augment it with a fixed module for multi-scale feature aggregation. DPC employs a simple blackbox optimization strategy, namely a combination of random search and local search, while Auto-DeepLab leverages a one-shot model and gradient-based NAS~\citep{darts}. Right: summary of (i) the different training phases (pretraining, architecture search, re-ranking, re-training and final evaluation), (ii) non-searchable components in each stage, and (iii) parameters that are optimized in each stage (weights associated with the non-searchable architectural component $w_{ns}$, weights associated with the searchable architectural component $w_{s}$, searchable architectural components $\alpha_s$).
		}
		\label{fig:comparison}
\end{figure*}

Unfortunately, manually designing neural network architectures comes with some major drawbacks, reminding of the drawbacks of manually designing features. Firstly, it is a time-consuming and error-prone process, requiring human expertise. This dramatically limits access to deep learning technologies since architecture engineering expertise is rare. Secondly, performance will be limited by the human imagination. 
Inspired by learning features from data rather than manually designing them, it seems natural to also replace the manual architecture design by learning architectures from data. This process of automating architectural engineering is commonly referred to as \emph{neural architecture search} (NAS). 

Until recently, NAS research has mostly focused on image classification problems, such as CIFAR-10 or ImageNet, due to the demand for computational resources in the order of hundreds or thousands of GPU days that early methods required~\citep{Zoph16,zoph-arXiv18,real_regularized_2018}. Compared to image classification, dense prediction tasks have barely been addressed even though they are of high practical relevance for applications, such as autonomous driving~\citep{huang2020autonomous} or medical imaging~\citep{LITJENS201760}. These problems are intrinsically harder than image classification for several reasons: they typically come with longer training times as well as higher memory footprints due to high-resolution data, and they also require more complex neural architectures. These differences lead to even higher computational demands and make the application of many NAS approaches problematic. Early works on NAS for dense prediction tasks (e.g., \citet{NIPS2018_8087, Ghiasi_2019_CVPR}) are thus limited to optimizing only small parts of the overall architectures, while still requiring enormous computational resources even though employing various tricks for speeding up the search process.

Fortunately, recent weight-sharing approaches via one-shot-models~\citep{SaxenaV16,bender_icml:2018,Pham18, darts, cai2018proxylessnas,Xie18} have dramatically reduced the computational costs to essentially the same order of magnitude as training a single network, making NAS applicable to a much wider range of problems. This lead to an increasing interest in developing NAS approaches tailored towards dense prediction tasks. However, due to the complex nature of the problem, these approaches vary vastly, as illustrated in Figure~\ref{fig:comparison}. With this survey, we aim to provide guidance to the most important design decisions.

This manuscript is structured as follows: in Section~\ref{sec:recap}, we briefly review NAS. We then discuss NAS for dense prediction tasks in general in Section~\ref{sec:nas_dpt}. In the remaining sections, we focus on the specific problems of semantic segmentation (Section~\ref{sec:ss}) and object detection (Section~\ref{sec:od}) and conclude by discussing other less-studied but promising applications (Section~\ref{sec:misc}).

%% file: chapters/recap_and_nas_for_dpt.tex
\section{A Brief Recap of NAS}
\label{sec:recap}

We briefly review neural architecture search; please refer to the surveys by \citet{elsken_survey} or \citet{wistuba2019survey} for a more thorough overview. 

\noindent{}\emph{Neural architecture search} (NAS) is typically framed as a bi-level optimization problem
\begin{equation*} \label{eq:nas_bilevel}
\begin{split}
\min_{A \in \mathcal{A}} & \, \mathcal{L}_{val}\big(D_{val}, A, w^*_A \big)          \\
s.t. \quad & w_A^{*} \in {arg\,min}_{w}  \mathcal{L}_{train}(D_{train},A, w),
\end{split}
\end{equation*}

\begin{figure*}
		\centering
\includegraphics[width=\linewidth]{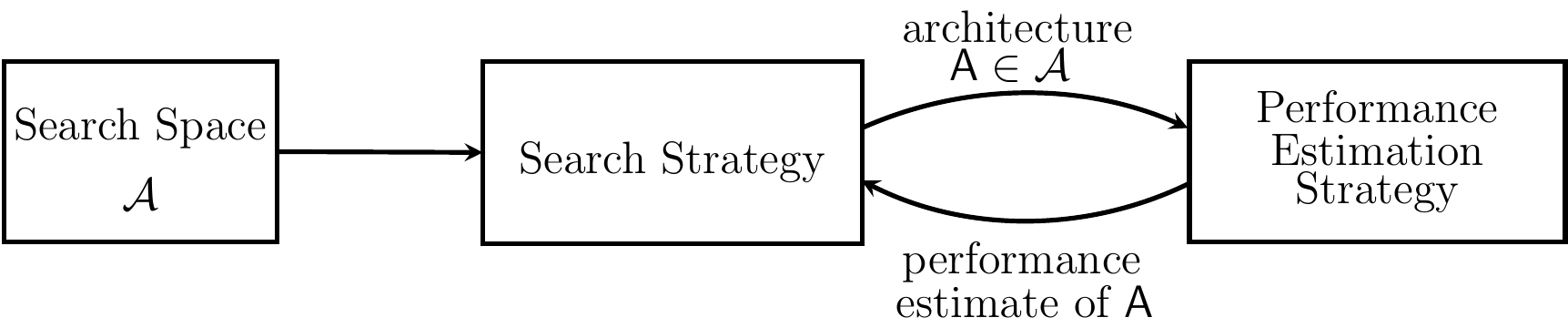}	
		\caption{
		Different dimensions of NAS algorithms. A search strategy selects an architecture from a predefined search space. The architecture is passed to a performance estimation strategy, which returns the estimated performance to the search strategy.
Taken from \citet{elsken_survey}.
		}
		\label{fig:nas_dimensions}
\end{figure*}

\noindent{}with the goal of finding an optimal neural network architecture $A$ within a search space $\mathcal{A}$ with respect to a validation loss function $\mathcal{L}_{val}$, a validation data set $D_{val}$ and weights $w^*_A$ of the architecture obtained by minimizing a training loss function $\mathcal{L}_{train}$ on a training data set $D_{train}$. NAS methods can be categorized along three dimensions~\citep{elsken_survey}: search space, search strategy and performance estimation, compare Figure \ref{fig:nas_dimensions}.

The \emph{search space} defines which architectures can be discovered in principle. Searchable components of an architecture can be architectural hyperparameters, such as the number of layers, the number of filters, or kernel sizes for convolutional layers, but also the layer types themselves, e.g., whether to use a convolutional or a pooling layer. Furthermore, NAS methods can optimize in which form layers are connected to each other, i.e., they search for the topology of the graph associated with a neural network. 

Building prior knowledge about neural network architectures into a search space can simplify the search. For instance, inspired by popular manually designed architectures, such as ResNet~\citep{he-cvpr16} or Inception-v4~\citep{Szegedy:2017:III:3298023.3298188}, \citet{zhong_practical_2017} and \citet{zoph-arXiv18} proposed to search for repeatable building blocks (referred to as cells) rather than the whole architecture. These building blocks are then simply stacked in a pre-defined manner to build the full model. 

Restricting the search space to repeating building blocks limits methods to only optimize these building blocks 
rather than also discovering novel connectivity patterns and ways of constructing architectures on a macro level from a set of building blocks. \citet{Yang2020NAS} show that the most commonly used search space is indeed very narrow in the sense that almost all architectures perform well. As a consequence, simple search methods, such as random search can be competitive~\citep{li_repro_nas,sciuto19,Elsken17}.
We note that this does not necessarily hold for richer, more diverse search spaces~\citep{Bender_2020_CVPR,real2020automl}. In contrast, one could also build as little prior knowledge as possible into the search space, e.g., by searching over elementary mathematical operations~\citep{real2020automl}, however, this would significantly increase the search cost. In general, there is typically a trade-off between search efficiency and the diversity of the search space.

Common \emph{search strategies} used to find an optimal architecture within a search space are black-box optimizers, such as evolutionary algorithms~\citep{stanley_evolving_2002, Real17,liu-iclr18,real_regularized_2018,Elsken19}, reinforcement learning~\citep{Zoph16, baker-iclr17,zhong_practical_2017, zoph-arXiv18} or Bayesian optimization~\citep{SweDuvSnoHutOsb13,mendoza_towards_2016,kandasamy_neural_2018,NIPS2019_8557,white2019bananas,ru_nasbowl}. As these methods typically require training hundreds or thousands of architectures and thus result in high computational costs, several newer methods tailored towards NAS go beyond this blackbox view. A popular approach to speed up this search is to employ a continuous relaxation of the architecture search space~\citep{darts}, which also allows for gradient-based optimization. In this line of research, rather than making a discrete decision for choosing one out of many candidate operations (such as convolution or pooling), a weighted sum of candidates is used, whereas the weights can then be interpreted as a parameterization of the architecture.

The objective function to be optimized by NAS methods is typically the performance an architecture would obtain after running a predefined (or also optimized) training procedure. However, this true performance is typically too expensive to evaluate. Therefore, various methods for \emph{estimating the performance} have been developed. A common strategy to speed up training is to employ lower-fidelity estimates (e.g., training for fewer epochs, training on subsets of data or downscaled images, and using downscaled architectures in the search phase~\citep{Chrabaszcz_arXiv17,baker_accelerating_2017,zoph-arXiv18,Zela18,Zhou_2020_CVPR}). Another popular approach is to employ weight sharing between architectures within one-shot-models~\citep{SaxenaV16,bender_icml:2018,Pham18,darts} as this  overcomes the need for training thousands of architectures. Rather than considering different architectures independently of each other, a single one-shot model is built to subsume all possible elements of the search space. Individual architectures are then simply subgraphs of the one-shot model and the weights of the one-shot model are shared across subgraphs. Another line of works focuses on \emph{predicting} the performance of neural network architectures, e.g., via trainable surrogate models~\citep{Wen2020NeuralPF,siems20,NEURIPS2020_768e7802}, considering learning curves~\citep{DomSprHut15,ru2021revisiting,baker_accelerating_2017,klein-iclr17} or zero-cost methods that are typically based on the statistics of an architecture or a single forward pass through the architecture~\cite{mellor2021neural,lee2018snip,abdelfattah2021zerocost}. We refer the interested reader to \citet{White2021HowPA} for a recent overview and comparison of such approaches.

\paragraph{Hardware-awareness.} Recently, many researchers also consider the resource consumption of neural networks, e.g., in terms of latency, model size, or energy consumption as objectives in NAS, since these are severely limited in many applications of deep learning. The importance of this fact is reflected by a whole line of research on manually designing top-performing yet resource-efficient architectures \citep{iandola_squeezenet:_2016,howard_mobilenets:_2017,Sand18,Zhang_2018_CVPR,Ma_2018_ECCV, Gholami_2018_CVPR_Workshops}. Many NAS methods also consider such requirements for dense prediction tasks by now, e.g., \citet{Zhang_2019_CVPR, Liu_2019_CVPR,Shaw_2019_ICCV,lin_2020_cvpr,Li_2020_CVPR,Chen_2020_ICLR,Bender_2020_CVPR,Guo_2020_CVPR,Chen_2020_CVPR}. Typically this is achieved by either adding a regularizer penalizing excessive resource consumption to the objective function~\citep{cai2018proxylessnas,mnasnet_tan} or by multi-objective optimization~\citep{Elsken19,10.1145/3321707.3321729}. We refer the interested reader to \citet{benmeziane2021comprehensive} for a more general discussion on this topic.

\section{NAS for Dense Prediction Tasks}
\label{sec:nas_dpt}

\begin{figure*}
		\centering
\includegraphics[width=\linewidth]{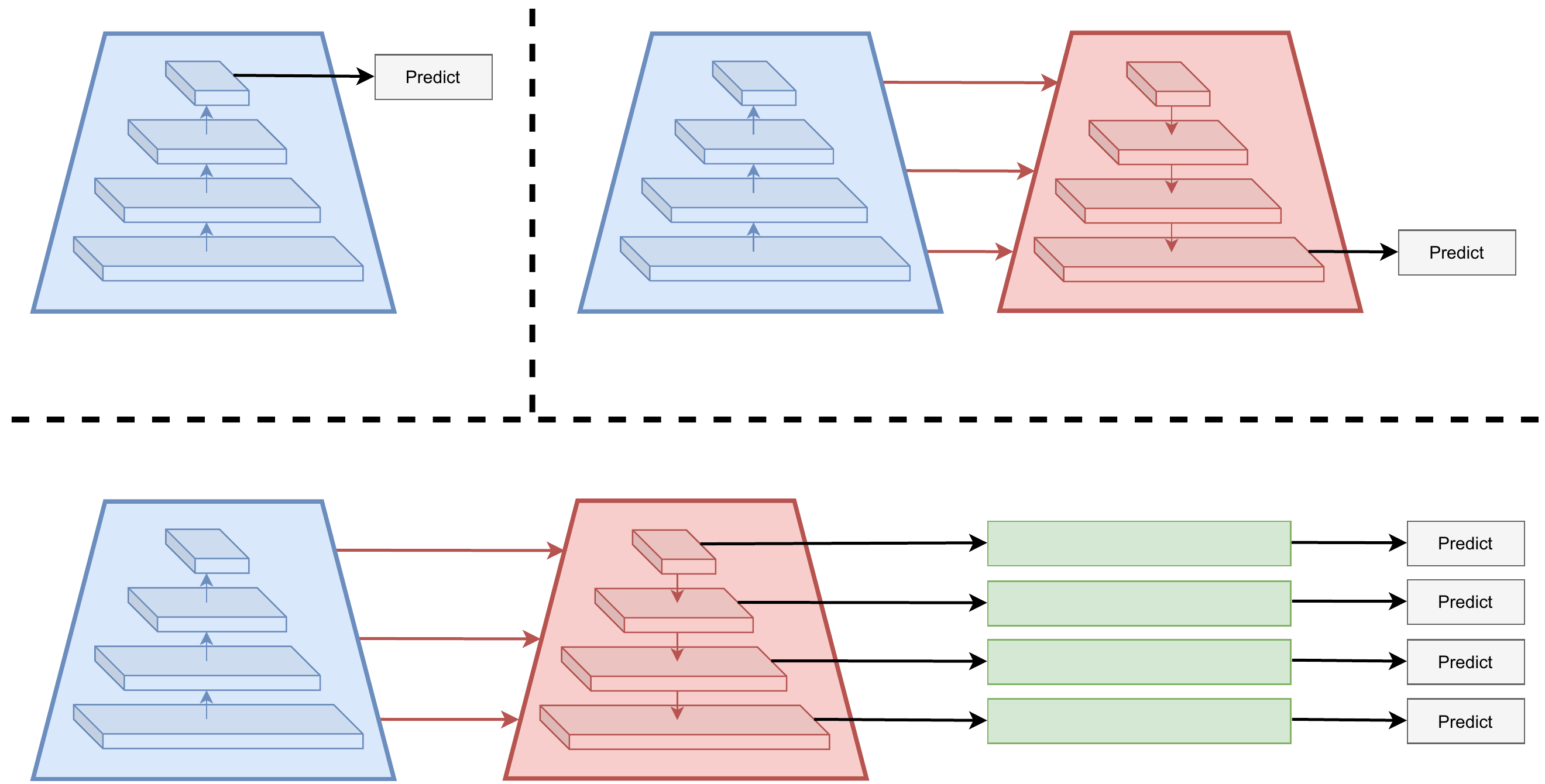}	
		\caption{High-level illustration of architectures employed for different tasks. Top left: typical encoder-like architecture (blue) for image classification problems; predictions are made based on low-resolution but semantically strong features. Top right: typical architecture for tasks like semantic segmentation; semantically strong features are generated for all scales through augmenting the encoder with a decoder (red). Bottom: semantically strong features from all scales serve as the input for the object detection head (green); note that the feature maps within the encoder and decoder might be densely connected and feature maps in the decoder might be connected to any other feature map in the encoder as well as the decoder. }
		\label{fig:high_level_archs}
\end{figure*}

Before discussing specific tasks in later sections, we first look at the aforementioned dimensions (search space, search strategy, and performance estimation) more generally in the context of dense prediction tasks, which come with novel challenges. For instance, compared to image classification problems, where one typically searches for encoder-like architectures, dense prediction tasks usually require much more complex architectures, e.g., to generate multi-scale features (which have been shown to be helpful for dense prediction tasks~\citep{Lin_2017_CVPR}).  We refer to Figure~\ref{fig:high_level_archs} for an illustrative comparison of commonly employed high-level architectures for image classification, semantic segmentation and object detection. Training models for dense prediction tasks is typically also much more demanding than for image classification, for at least three reasons: firstly, the spatial resolution is often higher, e.g., $32\times32$ for CIFAR-10~\citep{Krizhevsky09learningmultiple} or $224\times224$ for ImageNet~\citep{ILSVRC15} compared to $1024\times2048$ for Cityscapes~\citep{Cordts2016Cityscapes}. Secondly, the network's output is considerably larger as it requires a per-pixel prediction rather than a single prediction for a whole image. Consequently, the employed neural networks also tend to be much bigger. Both of these reasons lead to higher computational costs for training, as well as higher memory footprints. Lastly, data sets are typically smaller due to the higher annotation effort. Consequently, networks are often pretrained on other data sets to increase performance, again resulting in increased computational costs and complexity of the training pipeline.

 \paragraph{Search space.}
Due to the increased architectural complexity, many researchers focus on optimizing one component of the architecture for simplicity and efficiency. For the encoder (commonly referred to as the \emph{backbone}), the search spaces are often similar to search spaces for image classification, but more complex architectural building blocks are used. For image classification, these blocks are typically elementary operations, such as convolution or pooling layers. On the other hand, for dense prediction tasks, it is common to employ already pre-optimized blocks from state-of-the-art image classification networks~\citep{Shaw_2019_ICCV,Wu_2019_CVPR,Bender_2020_CVPR,NIPS2019_8890,Guo_2020_CVPR}, such as MobileNetV2~\citep{Sand18}, MobileNetV3~\citep{Howard_2019_ICCV} or ShuffleNetV2~\citep{Ma_2018_ECCV} and solely search over their architectural hyperparameters (e.g., the kernel size or the number of filters).
For other parts of the architectures, search spaces are typically built around well-performing manually designed architectures. For example, \citet{NIPS2018_8087} search for a dense prediction cell inspired by operations from DeepLab~\citep{deeplabv3,deeplab} and PSPNet~\citep{PSPNet}, \citet{Xu_2019_ICCV} build their space to contain FPN~\citep{Lin_2017_CVPR} and PANet~\citep{Liu_2018_CVPR}, and the space of \citet{Liu_2019_CVPR} contains the architectures proposed by \citet{7410535}, \citet{10.1007/978-3-319-46484-8_29} and \citet{deeplabv3}.

\paragraph{Performance estimation} plays an important role in making the search costs feasible. 
While lower-fidelity estimates -- such as conducting the search on downscaled models and training for fewer iterations -- are employed just like in NAS for image classification, NAS for dense prediction tasks also saves computational costs in other ways. Common approaches include employing pretrained models~\citep{NIPS2018_8087,Nekrasov_2019_CVPR,Guo_2020_CVPR,Wang_2020_CVPR,Chen_2020_CVPR}, caching of features generated by a backbone~\citep{NIPS2018_8087, Wang_2020_CVPR,Nekrasov_2019_CVPR}, or using not just a smaller but potentially also different backbone architecture~\citep{NIPS2018_8087, Ghiasi_2019_CVPR} in the search process. Lower-fidelity estimates are often used in multiple search phases. 
In the first stage, architectures are screened in a setting where they are cheap to evaluate (e.g., by using the tricks discussed above). Once a pool of well-performing architectures or candidate operations is identified, this pool is re-evaluated in a setting closer to the target setting (e.g., by scaling the model up to the target size or by training for more iterations). For example, \citet{NIPS2018_8087} explore 28,000 architectures in  a first stage with a downscaled and pretrained backbone, which is frozen during the search. The authors then choose the top 50 architectures found and train all of them fully to convergence. Rather than selecting top-performing architectures, \citet{Guo_2020_CVPR} propose a sequential screening of the search space to identify and remove poorly performing operations from the search space. All components of the architecture can then be jointly optimized on the reduced search space, which would have been infeasible on the full space due to memory limitations.

\paragraph{Search strategies} employed for dense prediction tasks are often build upon image classification methods. For example, many methods~\citep{Xu_2019_ICCV,Liu_2019_CVPR,Saikia_2019_ICCV,Zhang_2019_CVPR,Guo_2020_CVPR} use gradient-based techniques such as DARTS~\citep{darts}, often in its first-order approximation for computational reasons. Reinforcement learning based approaches are also re-used \citep{Ghiasi_2019_CVPR,Chen_2020_CVPR,Du_2020_CVPR,Wang_2020_CVPR,Bender_2020_CVPR}, and \citet{NIPS2019_8890} employ an evolutionary algorithm in combination with a one-shot model, as proposed by \citet{guo2019single} for image classification.

%% file: chapters/segmentation.tex
\section{Semantic Image Segmentation}
\label{sec:ss}

\subsection{Design Principles}
Semantic segmentation refers to the task of assigning a class label to each pixel of an image. 
The semantic segmentation model is trained to learn a mapping $f: \mathbb{R}^{w\times h\times c} \mapsto \mathbb{P}^{w\times h}$, where $w\times h$ refers to the spatial resolution, $c$ to the number of input channels, and $\mathbb{P}=\{(p_0,\dots,p_{C-1}) ~ \vert ~ p_i \in [0, 1] \wedge \sum_{i=0}^{C-1} p_i = 1\}$, with $C$ being the number of classes. \citet{long_fcnet} proposed to address this problem with deep learning by adapting image classification networks to produce dense outputs with fully convolutional neural networks. Related tasks to which the NAS methods discussed below can be applied without considerable changes are instance segmentation, which requires segmenting each object instance, and panoptic segmentation, which unifies semantic and instance segmentation. Popular data sets for semantic segmentation include PASCAL VOC~\citep{Everingham15}, Cityscapes~\citep{Cordts2016Cityscapes}, ADE20K~\citep{zhou2016semantic,zhou2017scene}, CamVid~\citep{BrostowFC:PRL2008,BrostowSFC:ECCV08}, and MS COCO~\citep{10.1007/978-3-319-10602-1_48}.

Several years of manual neural architecture engineering for semantic segmentation have identified several concepts that can be used when designing a search space for NAS: 
\\ \textit{1. Encoder-decoder macro-architecture~\citep{long_fcnet,unet}}: while input and output of a semantic segmentation model have the same spatial resolution, addressing the ``what is in an image?''  question typically requires integrating long-range spatial dependencies in the input. As this becomes easier on downsampled representations of the input, a popular approach is to use an encoder that gradually decreases spatial resolution while generating more abstract representations of the input. Using such a scale-decreased encoder has the additional advantages of being more computationally efficient and being able to use adapted feature extractors that were pretrained for image classification on ImageNet as encoders. Answering the ``where?'' question at full spatial resolution is the task of the decoder, which learns to gradually upsample the lower-resolution output of the encoder to the full image resolution.
\\ \textit{2. Skip connections~\citep{unet}}: while the encoder-decoder macro-architecture is efficient in addressing the ``what?'' question, the low-resolution bottleneck between encoder and decoder loses spatial precision which makes it unnecessarily difficult to adequately answer the ``where?'' question. One way of addressing this is to add higher resolution skip connections between encoder and decoder that bypass the bottleneck. The popular U-Net architecture~\citep{unet} introduces these skip connections between identical spatial resolutions of the encoder and the decoder in order to avoid any loss of detail through the bottleneck.
\\ \textit{3. Common building blocks}: building blocks used in neural architectures for image classification, such as residual or dense blocks, can be readily re-used in the encoder architecture. Moreover, search spaces for neural cells for image classification can also be utilized when applying NAS to the encoder part of semantic segmentation.
\\ \textit{4. Multi-scale integration}: augmenting encoder-decoder-like macro-architectures with a specific component that supports multi-scale integration helps capturing long-range dependencies. (Atrous) spatial pyramid pooling (ASPP)~\citep{7005506,deeplab} is one popular approach to this.

A NAS search space for semantic segmentation based on these design principles can thus learn 
\begin{enumerate}[label=(\alph*)]
    \item building blocks/cells used in the encoder,
    \item the downsampling strategy of the encoder,
    \item the building blocks/cells of the decoder,
    \item the upsampling strategy of the decoder,
    \item where and how to add skip connections between encoder and decoder, and
    \item how to perform multi-scale integration.
\end{enumerate}

Learning only (a) and/or (c) would be similar to the so-called \textit{micro} search, since the backbone/decoder is fixed and the NAS algorithm only searches for the optimal structure of the building blocks. On the other hand, we shall refer as \textit{macro} search to approaches that optimize for at least one of the other components besides (a) and/or (c). While the components above are the canonical components to be optimized, we would also like to note that a promising direction for future work on NAS for semantic segmentation is to define search spaces that allow exploring architectures that do not follow the predominant encoder-decoder design principle~\citep{Du_2020_CVPR}.

\subsection{NAS for Semantic Segmentation}

We refer to Table~\ref{tbl:ss} for an overview and comparison of the methods that we discuss in the following. The table is structured according to the criteria discussed in Section~\ref{sec:nas_dpt}.

\begin{table*}
\resizebox{1.0\linewidth}{!}{
\begin{tabular}{cccccccccccc}
\multirow{2}{*}{\textbf{method}} & \multicolumn{2}{c}{\textbf{search space}}                                                                                                             & \multirow{2}{*}{\textbf{\begin{tabular}[c]{@{}c@{}}search \\ method\end{tabular}}}& \multicolumn{2}{c}{\textbf{performance estimation}}                                                                                         & \multirow{2}{*}{\textbf{\begin{tabular}[c]{@{}c@{}}resource\\ efficiency \\ considered\end{tabular}}} & \multirow{2}{*}{\textbf{\begin{tabular}[c]{@{}c@{}}search \\ costs\end{tabular}}} & \multicolumn{2}{c}{\textbf{data sets}}                                                                           \\ \cline{2-3} \cline{5-6} \cline{9-10} 
         &                                 backbone & \begin{tabular}[c]{@{}c@{}}multi-level\\ features extractor\end{tabular} &                                         & \begin{tabular}[c]{@{}c@{}}weight\\ sharing\end{tabular} & 
         \begin{tabular}[c]{@{}c@{}}pre-\\training\end{tabular} &                                                                                                       &                                                                                   & \begin{tabular}[c]{@{}c@{}}searched\\ on\end{tabular} & \begin{tabular}[c]{@{}c@{}}transferred\\ to\end{tabular} \\ \hline \hline

         \begin{tabular}[c]{@{}c@{}} Auto-DeepLab\\ \citep{Liu_2019_CVPR} \end{tabular}  
    
          & \cmark 
          &  \xmark  
          & GB
          & \cmark
          & \xmark
          & \xmark 
          & small
          &  Cityscapes
          &\begin{tabular}[c]{@{}c@{}} PASCAL VOC,\\ ADE20K \end{tabular}   \\  \hline

          \begin{tabular}[c]{@{}c@{}} SqueezeNAS\\ \citep{Shaw_2019_ICCV} \end{tabular}  
    
          & \cmark 
          &  \xmark  
          & GB
          & \cmark
          & \xmark 
          & \cmark 
          & small
          &  Cityscapes
          & \\          \hline
          
          \begin{tabular}[c]{@{}c@{}} CAS\\ \citep{Zhang_2019_CVPR} \end{tabular}  
          & \cmark 
          &  \cmark  
          & GB
          & \cmark
          & \cmark
          & \cmark 
          &  small
          &  Cityscapes
          & CamVid\\  \hline
          
          \begin{tabular}[c]{@{}c@{}} FNA \\ \citep{Fang2020Fast} \end{tabular}  
          & \cmark
          & \xmark  
          & GB 
          & \cmark
          & \cmark
          & \cmark
          &  small 
          &  Cityscapes
          &  \\      \hline
          
          \begin{tabular}[c]{@{}c@{}} DPC\\ \citep{NIPS2018_8087} \end{tabular}  
          & \xmark 
          &  \cmark  
          & RS+LS
          & \xmark
          & \cmark 
          & \xmark
          &  high
          &  Cityscapes
          & Pascal VOC\\ \hline
          
          \begin{tabular}[c]{@{}c@{}} FasterSeg\\ \citep{Chen_2020_ICLR} \end{tabular}
          & \cmark
          & \xmark
          & GB
          & \cmark
          & \xmark
          & \cmark
          & small
          & Cityscapes
          & CamVid, BDD \\  \hline
          
          \begin{tabular}[c]{@{}c@{}} GAS\\ \citep{lin_2020_cvpr} \end{tabular}
          & \cmark
          & \xmark
          & GB
          & \cmark
          & \xmark
          & \cmark
          & small
          & Cityscapes
          & CamVid  \\  \hline
          
          \begin{tabular}[c]{@{}c@{}} \citet{Nekrasov_2019_CVPR} \end{tabular}
          & \xmark
          & \cmark
          & RL
          & \xmark
          & \cmark
          & \xmark
          & small
          & \begin{tabular}[c]{@{}c@{}} PASCAL VOC, \\ BSD, MS COCO \end{tabular}
          & \begin{tabular}[c]{@{}c@{}}MS COCO,\\ MPII, NYUDv2  \end{tabular}\\ \hline
          
          \begin{tabular}[c]{@{}c@{}} SparseMask \\ \citet{Wu_2019_ICCV} \end{tabular}
          & \xmark
          & \cmark
          & GB
          & \xmark
          & \cmark
          & \cmark
          & small
          & PASCAL VOC
          & ADE20K, MSRA-B, BSD  \\ \hline
          
          \begin{tabular}[c]{@{}c@{}} DCNAS \\ \citet{zhang2021dcnas} \end{tabular}
          & \cmark
          & \xmark
          & GB
          & \cmark
          & \xmark
          & \xmark
          & small
          & Cityscapes
          & \begin{tabular}[c]{@{}c@{}}PASCAL VOC, ADE20K, \\ PASCAL-Context \end{tabular}\\ \hline
          
          \begin{tabular}[c]{@{}c@{}} NAS-Unet\\ \citet{Weng_2019_ieee} \end{tabular}
          & \cmark
          & \xmark
          & GB
          & \cmark
          & \xmark
          & \xmark
          & small
          & PASCAL VOC
          &  \begin{tabular}[c]{@{}c@{}}Promise12, \\ Chaos, NERVE \end{tabular} \\ \hline
          
          \begin{tabular}[c]{@{}c@{}} C2FNAS \\ \citet{Yu_2020_CVPR} \end{tabular}
          & \cmark
          & \xmark
          & EA+RS
          & \cmark
          & \xmark
          & \xmark
          & high
          & MSD Pancreas
          & MSD 10 \\ \hline
          
          \begin{tabular}[c]{@{}c@{}} V-NAS\\ \citet{Zhu_2019_3dv} \end{tabular}
          & \cmark
          & \xmark
          & GB
          & \cmark
          & \xmark
          & \xmark
          & small
          & NIH Pancreas
          & MSD Lung, Pancreas \\
\hline
\end{tabular}
}
\caption{Overview of different NAS methods for semantic segmentation. For search methods, EA, LS, GB, RL, and RS refer to evolutionary algorithm, local search, gradient-based, reinforcement learning and random search, respectively. Weight sharing refers to weight sharing via one-shot models~\citep{bender_icml:2018,Pham18}. Pretraining refers to ImageNet pretraining. Since the search costs depend on the hardware and are also not explicitly mentioned in each paper, we only categorize them as ``small'' and ``high''. We assign the cost label ``small'' to methods that can be run within a week on a server with eight GPUs, i.e., in less than $56$ GPU days. Methods with ``high'' search costs typically employ a large-scale, distributed infrastructure, resulting in hundreds or thousands of GPU/TPU days of compute.
}
\label{tbl:ss}
\end{table*}

\paragraph{Backbone Search.}

Auto-DeepLab~\citep{Liu_2019_CVPR} builds upon the DARTS~\citep{darts} search space and algorithm, which were initially designed for learning optimal convolutional cells for image classification. The authors extend DARTS to the semantic segmentation task by also considering the \textit{macro} architecture, in the sense that it does not search only for an optimal cell structure, but also searches for the optimal spatial resolution of the feature maps that each cell processes. More specifically, a cell $C^{l,s}$ at layer $l$ that outputs a tensor with spatial resolution $s$, can learn to process input tensors from previous layers with output tensors with resolutions $s/2$, $s$ or $2s$. This is performed by continuously relaxing these discrete choices as it is done for the operation choices inside the cells. This results in multiple network outputs, each having a different spatial resolution. Each of these outputs is connected with an ASPP module~\citep{deeplab}.
For optimizing the architectural weights (of both cell and macro architecture), the authors utilize first-order DARTS. 

A line of follow-up work improves Auto-DeepLab in various directions. \citet{Chen_2020_ICLR}, \citet{Shaw_2019_ICCV}, and \citet{lin_2020_cvpr} search for efficient architectures (e.g., by means of latency) by adding a regularizer for hardware-costs. Various more powerful search spaces are also proposed, e.g., to cover channel expansion ratios and multi-branch architectures~\citep{Chen_2020_ICLR}, or by employing stronger building blocks such as inverted residual blocks~\citep{Shaw_2019_ICCV} rather than simple convolutions, as well as removing the typical constraint that the cell topology is shared across the whole architecture~\citep{lin_2020_cvpr}. \citet{zhang2021dcnas} address DARTS' (and therefore also Auto-DeepLab's) problem of keeping the entire one-shot model in memory; this is done by sampling paths in the one-shot model rather than training the entire model at once, similar to the approaches by \citet{Xie18} and \citet{dong2019search}. Due to the memory efficiency, the search is directly conducted on the target space and data set rather than employing a proxy task.

Semantic segmentation is in particular important for medical image analysis~\citep{unet} and consequently, NAS methods are also applied to optimize on medical image data sets. NAS-Unet~\citep{Weng_2019_ieee} employs ProxylessNAS~\citep{cai2018proxylessnas} to automatically search for a set of downsampling and upsampling cells that are connected using a Unet-like~\citep{unet} backbone.
\citet{Yu_2020_CVPR} and \citet{Zhu_2019_3dv} consider 3D medical image segmentation.
For this task, Coarse-to-Fine NAS (C2FNAS)~\citep{Yu_2020_CVPR} uses a search space inspired by the one employed in Auto-DeepLab~\citep{Liu_2019_CVPR} and an evolutionary strategy operating on clusters of similar networks to search for the macro structure of their model, whilst the operation choices inside the cells of the macro structure are randomly sampled similarly to the protocol by \citet{li_repro_nas}. Finally, V-NAS~\citep{Zhu_2019_3dv} extends DARTS to encoder-decoder architectures used for volumetric medical image segmentation.

\paragraph{Multi-Scale Feature Search.}
\citet{NIPS2018_8087} employ NAS for dense prediction tasks in order to search for a better multi-scale feature extractor called dense prediction cell (DPC) given a fixed backbone network. The proposed search space is a micro search space that contains, e.g., atrous separable convolutions with different rates or average spatial pyramid pooling inspired by DeepLabv3~\citep{deeplab}.
They run a combination of random search and local search to optimize the dense prediction cell given a fixed, pretrained backbone, which, despite the use of a series of proxy tasks, still required 2600 GPU days.

\citet{Nekrasov_2019_CVPR} also consider a fixed encoder network and search for an optimal decoder architecture together with the respective connections to the encoder layers. The decoder architecture is modeled as a sequence of cells sharing the same structure that processes the inputs from the encoder layers. 
The authors utilize various heuristics to speed-up architecture search. For example, they freeze the weights of the encoder network and train only the decoder part (as already done in DPC) and early-stop training of architectures with poor performance. Moreover, a knowledge distillation loss~\citep{distillation} is employed as well as an auxiliary cell to reduce the training time. Rather than using random and local search, a controller trained with reinforcement learning is employed to sample candidate architectures, similar to \citet{zoph-arXiv18}.

In follow-up work, \citet{nekrasov_2020_wacv} extend their work to semantic video segmentation by learning a \textit{dynamic cell} that learns to aggregate the information coming from previous and current frames to output segmentation masks.

\paragraph{Joint Search and Novel Design Principles.}

While previous work considers optimizing either the encoder or decoder, Customizable Architecture Search (CAS)~\citep{Zhang_2019_CVPR} searches for both an optimal backbone and multi-scale feature extractor, however in a sequential manner.
For the backbone, a normal cell (which preserves the spatial resolution and number of feature maps) and a reduction cell (which reduces the spatial resolution and increases the number of feature maps) are optimized. Once these two cells have been determined, a multi-scale cell is optimized to learn how to integrate spatial information from the backbone.

Rather than searching for optimized building blocks for the encoder and/or the decoder, \citet{Wu_2019_ICCV} propose to search for the connectivity pattern between the two components, which is typically fixed in other work. The encoder and decoder are first densely connected, where each connection is weighted by a real-valued parameter. This real-valued parameterization of the connections allows for gradient-based optimization as in DARTS~\citep{darts}. The authors also propose a loss function for inducing a sparse connectivity pattern.

Closely related to NAS techniques, \citet{Li_2020_CVPR} extend Auto-DeepLab by considering a differentiable gating function that learns data-dependent routes that propagate information at different scales depending on the input image. Moreover, they also consider budget constraints in their objective and use a one-shot model to find routing schemes that require fewer FLOPs compared to Auto-DeepLab.

%% file: chapters/object_det.tex
\section{Object Detection}
\label{sec:od}

\subsection{Design Principles}
Object detection \citep{od_survey} refers to the task of identifying if/how many objects of predetermined categories are present in an input (e.g., an image) and, for each identified object, determining its category as well as its spatial localization. Spatial localization can be represented in different ways, with the most common one being a 2D bounding box in image space, encoded by a 4D real-valued vector. However, other representations, such as pixel-wise segmentation, are possible as well. 
We note that in contrast to semantic segmentation, deep learning-based object detection often has a post-processing step that maps from dense network outputs to a sparse set of object detections, e.g., using non-maximum suppression. However, this post-processing is typically fixed and not used during training; and thus also ignored during NAS (we note that applying NAS to this post-processing would be an interesting future direction). Moreover, deep learning-based object detection can be split into one-stage and two-stage approaches. Two-stage approaches first identify the presence and extent of an arbitrary object at a position and thereupon apply a region classifier to the identified object region to classify the category of the object and (optionally) refine its spatial localization. In contrast, single-stage approaches directly predict the presence of an object, its class, as well as its spatial localization in a single forward pass.

Since objects can have vastly different scales, typically multi-scale approaches are applied for single-stage object detection. This can be achieved by either attaching ``detection heads'' at layers of different spatial resolutions or by combining features of different layers; effectively, this results in certain network outputs (those corresponding to lower resolutions) specializing on larger objects and higher resolution outputs on smaller ones. 
In this case, the dense prediction task can be framed as $f: \mathbb{R}^{w\times h\times c} \mapsto [\mathbb{P}^{w\times h} \times \mathbb{R}^{w\times h\times b},
 \mathbb{P}^{w/2\times h/2} \times \mathbb{R}^{w/2\times h/2\times b}, \dots]$, where $w\times h$ refers to the spatial resolution, $c$ to the number of input channels, $b$ to the parameters encoding the spatial localization, and $\mathbb{P}=\{(p_{-1}, p_0,\dots,p_{C-1}) \vert p_i \in [0, 1] \wedge \sum_{i=-1}^C p_i = 1\}$, with $C$ being the number of classes and -1 corresponding to the "no object" class.

Many of the design principles of  semantic segmentation carry over to object detection. However, there are also notable differences:
\begin{itemize}
    \item Since the network requires a dense and multi-scale output, a further design choice is how ``detection heads'' generating these multi-scale outputs are attached to the main network. The heads' architecture itself is another open design choice.
   \item Two-stage object detection can impose complex interdependencies between the architectures of the two stages, making the design of a search space covering both stages together challenging.
\end{itemize}

\subsection{NAS for Object Detection}

We summarize the methods that we discuss in the following in Table~\ref{tbl:comp_od}. The table is again structured according to the criteria discussed in Section \ref{sec:nas_dpt}.

\begin{table*}[!ht]
\resizebox{1.0\linewidth}{!}{
\begin{tabular}{cccccccccccc}
\multirow{2}{*}{\textbf{method}} & \multicolumn{3}{c}{\textbf{search space}}                                                                                                             & \multirow{2}{*}{\textbf{\begin{tabular}[c]{@{}c@{}}search \\ method\end{tabular}}}& \multicolumn{2}{c}{\textbf{performance estimation}}                                                                                         & \multirow{2}{*}{\textbf{\begin{tabular}[c]{@{}c@{}}resource\\ efficiency \\ considered\end{tabular}}} & \multirow{2}{*}{\textbf{\begin{tabular}[c]{@{}c@{}}search \\ costs\end{tabular}}} & \multicolumn{2}{c}{\textbf{data sets}}                                                                           \\ \cline{2-4} \cline{6-7} \cline{10-11} 
                                 &                                 backbone & \begin{tabular}[c]{@{}c@{}}multi-level\\ features extractor\end{tabular} & \begin{tabular}[c]{@{}c@{}}task-specific\\ head(s)\end{tabular} &                                         & \begin{tabular}[c]{@{}c@{}}weight\\ sharing\end{tabular} & 
                    \begin{tabular}[c]{@{}c@{}}pre-\\training\end{tabular} &                                                                                                       &                                                                                   & \begin{tabular}[c]{@{}c@{}}searched\\ on\end{tabular} & \begin{tabular}[c]{@{}c@{}}transferred\\ to\end{tabular} \\ \hline \hline

          \begin{tabular}[c]{@{}c@{}} DetNAS\\ \citep{NIPS2019_8890} \end{tabular}  
          & \cmark
          &  \xmark  
          & \xmark  
          & EA 
          & \cmark 
          & \cmark   
          & \xmark
          &  small 
          & \begin{tabular}[c]{@{}c@{}}  Pascal VOC,\\MS COCO  \end{tabular}  
          &  \\ \hline

          \begin{tabular}[c]{@{}c@{}} NATS\\ \citep{NIPS2019_9576} \end{tabular}  
          & \cmark
          &  \xmark  
          & \xmark  
          & GB 
          & \cmark
          & \cmark
          &  \na
          & small 
          &  MS COCO
          &  \\ \hline
          
         \begin{tabular}[c]{@{}c@{}} TuNAS \\ \citep{Bender_2020_CVPR} \end{tabular}  
          & \cmark
          & \xmark  
          & \xmark  
          & RL
          & \cmark
          & \xmark
          & \cmark
          &  small
          &  MS COCO
          &  \\ \hline
          
          \begin{tabular}[c]{@{}c@{}} MobileDets \\ \citep{MobileDets} \end{tabular}  
          & \cmark
          & \xmark  
          & \xmark  
          & RL
          & \cmark
          & \xmark
          & \cmark
          &  small
          &  MS COCO
          &  \\ \hline
                             
         \begin{tabular}[c]{@{}c@{}} SP-NAS \\ \citep{Jiang_2020_CVPR} \end{tabular}  
          & \cmark
          & \xmark  
          & \xmark  
          & LS
          & \cmark
          & \cmark
          & \cmark
          &  small
          &  \begin{tabular}[c]{@{}c@{}} MS COCO, ECP,\\ PASCAL VOC, BDD \end{tabular}
          &  \\          \hline
          
         \begin{tabular}[c]{@{}c@{}} FNA \\ \citep{Fang2020Fast} \end{tabular}  
          & \cmark
          & \xmark  
          & \xmark  
          & GB
          & \cmark
          & \cmark
          & \cmark
          &  small
          &  MS COCO
          &  \\        \hline
                                 
        \begin{tabular}[c]{@{}c@{}}NAS-FPN \\ \citep{Ghiasi_2019_CVPR} \end{tabular}  
        & \xmark
        & \cmark
        & \xmark
        & RL 
        & \xmark 
        & \xmark
        & \xmark 
        & high
        &  MS COCO 
        &   \\ \hline
             
          \begin{tabular}[c]{@{}c@{}} MnasFPN\\ \citep{Chen_2020_CVPR} \end{tabular}  
          &  \xmark 
          & \cmark
          & \xmark  
          & RL
          & \xmark  
          & \cmark
          & \cmark  
          &  high
          &  MS COCO
          &     \\ \hline

          \begin{tabular}[c]{@{}c@{}} Auto-FPN\\ \citep{Xu_2019_ICCV} \end{tabular}  
          & \xmark 
          &  \cmark  
          & \cmark  
          & GB
          & \cmark
          & \cmark
          & \cmark
          & small
          & \begin{tabular}[c]{@{}c@{}}  Pascal VOC,\\MS COCO, BDD  \end{tabular}  
          & VG, ADE\\ \hline

         \begin{tabular}[c]{@{}c@{}} NAS-FCOS\\ \citep{Wang_2020_CVPR} \end{tabular}  
          & \xmark 
          &  \cmark  
          & \cmark
          & RL 
          &  \xmark  
          & \cmark
          &  \xmark
          & small
          & Pascal Voc
          & MS COCO \\ \hline

         \begin{tabular}[c]{@{}c@{}} FAD\\ \citep{zhong2020representation} \end{tabular}  
          & \xmark 
          &  \xmark  
          & \cmark
          & GB 
          &  \cmark  
          &  \na
          &  \xmark
          & small
          & Pascal Voc, MS COCO
          &  \\  \hline

          \begin{tabular}[c]{@{}c@{}} Hit-Detector\\ \citep{Guo_2020_CVPR} \end{tabular}  
          & \cmark
          & \cmark
          & \cmark  
          & GB 
          & \cmark
          &  \cmark
          & \cmark
          &  small
          &  MS COCO
          &  \\ \hline

        \begin{tabular}[c]{@{}c@{}} SM-NAS \\ \citep{yao_2020_aiii} \end{tabular}  
          & \cmark
          & \cmark
          & \cmark  
          & EA
          & \xmark  
           & \cmark
          &  \cmark
          &  high
          & MS COCO
          &   Pascal VOC, BDD \\ \hline

         \begin{tabular}[c]{@{}c@{}} SpineNet \\ \citep{Du_2020_CVPR} \end{tabular}  
          & \cmark
          & \cmark
          & \xmark  
          & RL
          & \xmark  
           & \xmark
          & \na 
          &  high
          & MS COCO
          &   \\

\hline
\end{tabular}
}
\caption{Overview of different NAS methods for object detection. For search methods, EA, LS, GB, and RL refer to evolutionary algorithm, local search, gradient-based and reinforcement learning, respectively. Weight sharing refers to weight sharing via one-shot models~\citep{bender_icml:2018,Pham18}. Pretraining refers to ImageNet pretraining. Since the search costs depend on the hardware and are also not explicitly mentioned in each paper, we only categorize them as ``small'' and ``high''. We assign the cost label ``small'' to methods that can be run within a week on a server with eight GPUs, i.e., in less than $56$ GPU days. Methods with ``high'' search costs typically employ a large-scale, distributed infrastructure, resulting in hundreds or thousands of GPU/TPU days of compute. 
}
\label{tbl:comp_od}
\end{table*}

Early work on NAS for object detection focuses on either optimizing the backbone or the multi-scale feature extractor. We start by discussing these two, orthogonal, directions and then go on to methods that jointly search all components, and to other search space design principles.

\paragraph{Backbone Search.} Since NAS is a technique that comes from image classification, considering that researchers typically employ image classification architectures as backbones for object detection, it is not surprising that a comprehensive line of research adapts existing methods for optimizing the backbone~\citep{NIPS2019_8890,Bender_2020_CVPR}. \citet{Bender_2020_CVPR} propose TuNAS, inspired by ProxylessNAS~\citep{cai2018proxylessnas} and ENAS~\citep{Pham18},  for image classification and also evaluate it on object detection, with only minor hyperparameter adjustments required. In contrast to most other work, \citet{Bender_2020_CVPR} train a one-shot model from scratch directly on the target task rather than employing pretraining. To improve the scalability with respect to the search space, the authors propose a more aggressive weight sharing across candidate choices, e.g., by sharing filters weights across convolutions with a different number of filters. Furthermore, the memory footprint when training the one-shot model is dramatically reduced by ``rematerialization'', i.e., re-computing intermediate activations rather than storing them. The authors also propose a novel hardware regularizer allowing to find models closer to the desired hardware cost. In a follow-up work~\citep{MobileDets}, the performance of TuNAS is further improved due to a more powerful search space.

Rather than searching for an architecture from scratch, \citet{NIPS2019_9576} propose to transform a given, well-performing backbone. Motivated by improving the effective receptive field size of convolutions, the authors search over various dilation rates. For each dilation rate, channels are grouped to allow for different dilation rates for different groups. The parameters of convolutional kernels are inherited from a pretrained model and shared across all rates to avoid additional parameters. Gradient-based architecture search on a continuous relaxation of the search space is used to then search for the optimal dilation rates for each channel group.
In a similar fashion, \citet{Jiang_2020_CVPR} modify an existing, well-performing and pretrained backbone by applying network morphisms~\citep{pmlr-v48-wei16}, which are commonly used in NAS~\citep{Elsken17,Elsken19,cai2017reinforcement,cai_path-level_2018}, to improve the backbone. Since network morphisms inherit the performance of the parent network to the child network, the child network does not need to be trained from scratch and thus the authors avoid pre-training all candidate architectures on ImageNet, which would be infeasible. In a first search phase, a purely sequential model is optimized, while a second search phase adds parallel branches to enable more powerful architectures. \citet{Fang2020Fast} also employ network morphisms, or more general parameter remapping methods, to initialize a one-shot model from a pretrained model to avoid pretraining the one-shot model.

\paragraph{Multi-Scale Feature and Head Search.}
Orthogonal to the methods discussed above, another line of research focuses on optimizing multi-scale feature extractors as well as the object detection head. \citet{Ghiasi_2019_CVPR} employ a reinforcement learning-based NAS framework~\citep{Zoph16,zoph-arXiv18} to search for NAS-FPN, an improved feature pyramid network (FPN)~\citep{Lin_2017_CVPR} yielding multi-scale features. 
In follow-up work, \citet{Chen_2020_CVPR} extend the search space and employ Mnas-Net~\citep{mnasnet_tan} as a search method for not only optimizing performance but also latency to find efficient networks. This is in contrast to NAS-FPN, where lightweight architectures are searched after manually adapting non-searchable components to be efficient. As both NAS-FPN and Mnas-FPN are based on expensive, black-box NAS methods that train around 10,000 architectures, they require substantial computational resources to be run, in the order of hundreds of TPU days. 

While the aforementioned work focuses on the FPN module, \citet{Wang_2020_CVPR} additionally optimize the object detection head on top of the multi-scale features. This is done by using RL to first search for an FPN-like module and afterwards for a detection head. For the FPN module, similar to \citet{Ghiasi_2019_CVPR}, the RL controller chooses feature maps from a list of candidates, an elementary operation to process, and in which way to merge it with another candidate. 
Once an optimal FPN is found, it is used to search for a suitable head. While typically the weights of the head are shared across all levels of the feature pyramid, \citet{Wang_2020_CVPR} also search over an index indicating from where on to share weights, while all layers of the head architecture before the index can have different weights for different pyramid levels. As the backbone architecture is not optimized, they pre-compute the output features from the backbone to make the search more efficient.

\citet{Xu_2019_ICCV} propose Auto-FPN, a method for searching for a multi-scale feature extractor and a detection head, based on a continuous relaxation and gradient-based optimization as done by \citet{darts}. Again, a cell-based search space is used, for both components. The search is conducted in a sequential manner (i.e., the FPN is searched first and the head afterward) as DARTS. Since the employed search strategy requires to keep the whole one-shot model in memory, it does not allow for a joint optimization in the considered setting. \citet{zhong2020representation} also employ DARTS to search for a detection head. To mitigate memory problems, they propose a more efficient scheme for sharing representations across operations with different receptive field sizes by re-using intermediate representations.

\paragraph{Joint Search and Novel Design Principles.}

The works  discussed so far focus on optimizing different parts of object detection architectures, but they all employed some non-searchable components. Given enough data and compute power, optimizing all components jointly should in principle dominate this approach. To give a concrete example of interaction effects between architecture components: while NAS-FPN~\citep{Ghiasi_2019_CVPR} in combination with a ResNet-50~\citep{He16} outperforms the original FPN module when also combined with a ResNet-50 (suggesting that NAS-FPN yields richer multi-scale features) and the DetNAS~\citep{NIPS2019_8890} backbone in combination with FPN outperforms FPN in combination with the original ResNet-50 backbone (suggesting that DetNAS is a better backbone), \citet{Guo_2020_CVPR} showed that the combination of the DetNAS backbone and NAS-FPN yields worse performance than ResNet-50 in combination with NAS-FPN. 
Therefore, they propose to search for all three components jointly. The main concern with this approach is that it can easily get computationally infeasible. To overcome this problem, a hierarchical search is proposed. In the first search phase conducted on a small proxy task, a rich search space (build around FBNet~\citep{Wu_2019_CVPR}) for all three components is explored with the goal of shaping the search space by pruning building blocks that are unlikely to be optimal. Notably, this allows starting with the same set of candidate operations for all three components, while in prior work the set of candidates is typically adapted to the specific component to be optimized~\citep{Xu_2019_ICCV}. By imposing a regularizer enforcing sparsity among the architectural parameters in the one-shot model in the first phase, suboptimal candidates can naturally be pruned away. The second search phase uses the resulting pruned sub-space to determine an optimal architecture. Both search phases employ gradient-based optimization for efficiency. Furthermore, the authors penalize architectures with high computational costs by adding a proper regularization term. Similarly, \citet{yao_2020_aiii} first search for the best combination of backbone, multi-scale feature extractor, region-proposal network as well as detection head, with a set of possible candidates for each component (e.g. ResNet or MobileNet V2 as a backbone or different versions of FPNs for multi-scale feature fusion). In the second stage, the best-performing combinations of these components are then fine-tuned on a more fine-grained level, e.g., by optimizing the number of channels in the chosen backbone.

While all previously discussed work is guided by manually designed architectures that consist of a scale-decreasing backbone followed by multi-scale feature fusion, \citet{Du_2020_CVPR} question this design principle and propose to search for a single network covering both components. 
This approach permutes layers of the network and searches for a better connectivity pattern between them. We highlight that for this search space consisting of layer permutations it is unclear how one-shot models could be employed and thus the authors rely on the computationally more expensive black-box optimization via RL.

%% file: chapters/ongoing_future_work.tex
\section{Outlook: promising application domains and future work}
\label{sec:misc}

So far, most NAS research on dense prediction has focused on semantic segmentation or object detection, but there are many more dense prediction tasks domains where the discussed methods could be applied or adapted to. 

For example, \emph{disparity estimation} can be solved in an end-to-end fashion with encoder-decoder architectures~\citep{MIFDB16}. First studies in this direction have already been conducted. \citet{Saikia_2019_ICCV}
propose AutoDispNet, which extends the typical search space from image classification consisting of a normal and a reduction cell by an upsampling cell in order to search for encoder-decoder architectures. The first order approximation of DARTS is used to allow an efficient search, followed by a hyperparameter optimization for the discovered architectures using the popular multi-fidelity Bayesian optimization method BOHB~\citep{pmlr-v80-falkner18a}. \citet{NEURIPS2020_fc146be0} build upon AutoDispNet by also searching for a matching network on top of the feature extractor, inspired by recent manually designed networks for disparity estimation. 
Architectures discovered for disparity estimation~\citep{Saikia_2019_ICCV} or semantic segmentation~\citep{Nekrasov_2019_CVPR} have also been evaluated on \emph{depth estimation}.

\citet{Ulyanov_2018_CVPR} showed that the structure of an encoder-decoder architecture employed as a generative model is already sufficient to capture statistics of natural images without any training. Thus, such architectures can be seen as a ``deep image prior'' (DIP), which can be used to parameterize images. On a variety of tasks, such as \emph{image denoising, super-resolution or inpainting}, a natural image could successfully be generated from random noise and a randomly initialized encoder-decoder architecture. As the authors noted that the best results can be obtained by tuning the architecture for a particular task, \citet{ho2020neural} and \citet{NAS-DIP} employed NAS to search for deep image prior architectures via evolution and reinforcement learning, respectively. Differentiable architecture search has also been adapted for image denoising by \citet{NEURIPS2020_c6e81542}.

Other promising tasks are \emph{panoptic segmentation} with some first work by \citet{NEURIPS2020_ec1f7645} and 3D detection and segmentation~\citep{tang_3d_nas}. Finally, \emph{optical flow estimation}~\citep{flownetv1,flownetv2,Sun_2018_CVPR} is a problem that has not been considered by NAS researchers so far, and it is conceivable that NAS methods could further improve performance on this task.